\documentclass[12pt, onecolumn,journal]{IEEEtran}
\IEEEoverridecommandlockouts
\usepackage{mathrsfs}
\usepackage{amssymb}
\usepackage{amsmath}
\usepackage{amssymb}
\usepackage{amsthm}
\usepackage{multirow,makecell}
\usepackage{makecell}
\usepackage[table]{xcolor}
\usepackage{subfigure}
\usepackage{setspace}
\usepackage{stfloats}
\usepackage{graphicx}
\usepackage{epstopdf}
\usepackage{soul}
\soulregister\cite7
\soulregister\ref7
\usepackage{url}
\usepackage{color}
\usepackage{xcolor}

\ifCLASSINFOpdf \else \fi

\usepackage{algorithm}
\usepackage{algorithmic}

\usepackage{cite}

\begin{document}
	\begin{spacing}{1.2}

\title{Anti-collision Technologies for Unmanned Aerial Vehicles: Recent Advances and Future Trends}
\author{Zhiqing Wei,
		Zeyang Meng,
        Meichen Lai,
        Huici Wu,
        Jiarong Han,
        Zhiyong Feng
\thanks{
This work was supported in part by the Beijing Natural Science Foundation under Grant L192031, in part by the National Natural Science Foundation of China under Grant 61901051, and in part by Young Elite Scientists Sponsorship Program by CAST (YESS20200283).
		
Zhiqing Wei, Zeyang Meng, Meichen Lai, Jiarong Han and Zhiyong Feng are with Key Laboratory of Universal Wireless Communications, Ministry of Education, School of Information and Communication Engineering, Beijing University of Posts and Telecommunications, Beijing, 100876, China (e-mail: \{weizhiqing, mengzeyang, laimeichen, hjr19991020, fengzy\}@bupt.edu.cn).
		
H. Wu is with the National Engineering Lab for Mobile Network Technologies, Beijing University of Posts and Telecommunications (BUPT), Beijing 100876, China, and also with Peng Cheng Laboratory, Shenzhen, China. (e-mail: dailywu@bupt.edu.cn)
		
Correspondence authors: Zhiqing Wei, Zhiyong Feng.
}}
% (email: \{weizhiqing,\}@bupt.edu.cn).}}

%\markboth{IEEE ,~Vol.~, No.~, ~2011}%
%{Shell \MakeLowercase{\textit{et al.}}: Bare Demo of IEEEtran.cls for Journals}

\maketitle

\begin{abstract}
		
Unmanned aerial vehicles (UAVs) are widely applied in civil applications, such as disaster relief, agriculture and cargo transportation{, and so on}. With the massive number of UAV flight activities, the anti-collision technologies aiming to avoid the collisions between UAVs and other objects have attracted much attention. 
The anti-collision technologies are of vital importance to guarantee the survivability and safety of UAVs.
In this article, a comprehensive survey on UAV anti-collision technologies is presented.
We firstly introduce laws and regulations on UAV safety which prevent collision at the policy level.
Then, the process of anti-collision technologies is reviewed from three aspects, i.e., obstacle sensing, collision prediction, and collision avoidance.
We provide detailed survey and comparison of the methods of each aspect and analyze their pros and cons.
Besides, the future trends on UAV anti-collision technologies are presented from the perspective of fast obstacle sensing and fast wireless networking.
Finally, we summarize this article.

\end{abstract}

\IEEEpeerreviewmaketitle
\section*{Glossary}
\begin{table}[!h]
	
	\renewcommand{\arraystretch}{1.1} %looser
	\begin{center}
		
		\begin{tabular}{ll}
			
			2-D & Two-Dimensional.\\
			3-D & Three-Dimensional.\\
			6G & 6th Generation Mobile Networks.\\
			ES-A2C & Experience-shared Advantaged Actor - Critic.\\
			ACA & Ant Colony Algorithm.\\
			ACTS & Airman Certificate Testing Service.\\
			\multirow{2}[0]{0.2\linewidth}{ADS-B} & Automatic Dependent Surveillance - \\
			& Broadcast.\\
			ANN & Artificial Neural Network.\\
			ARE & Adaptive and Random Exploration.\\
			ATCT & Air Traffic Control Tower.\\

		\end{tabular}
	\end{center}
\end{table}
\begin{table}[!h]
	
	\renewcommand{\arraystretch}{1.1} %looser
	\begin{center}
		
		\begin{tabular}{ll}

			CAAC & Civil Aviation Administration of China.\\
				\multirow{2}[0]{0.2\linewidth}{CogMOR-MAC} & MAC Protocol under Multi-channel Opportu-\\
			& nistic Reservation based on Cognitive Radio.\\
			CNN & Convolutional Neural Network.\\
			DQN & Deep Q Network.\\
			DRL & Deep Reinforcement Learning.\\
			EM & Expectation-Maximization.\\
			FAA & Federal Aviation Administration.\\
			FLARM & Flight Alarm.\\
			\multirow{2}[0]{0.2\linewidth}{G-FCNN} & Generalized Fuzzy Competitive Neural \\
			& Network.\\
			HVS & Human Vision System.\\
			IR & Infrared Radiation.\\
			ISAC & Integrated Sensing and Communication.\\
			JRC & Joint Radar and Communication.\\
			JSC & Joint Sensing and Communication.\\
			LSTM & Long Short-Term Memory.\\
			LWIR & Long Wavelength Infrared.\\
			MAC & Multiple Access Control.\\
			MIMO & Multiple-Input and Multiple-Output.\\
			\multirow{2}[0]{0.2\linewidth}{MMAC-DA} & Multi-Channel MAC Protocol with\\
			& Directional Antennas.\\
			NAS & United States National Airspace System.\\
			NIR & Near Infrared.\\
			OS & Operating System.\\
			PH & Pythagorean-Hodograph.\\
			RCIS & Radar-Communication Integrated System.\\
			RNN & Recurrent Neural Network.\\
			RTT & Rapidly-exploring Random Tree.\\
			SAA & Simulated Annealing Algorithm.\\
			SAR & Synthetic Aperture Radar.\\
			STC & Space-Time Coded.\\
			TCAS & Traffic Collision Avoidance System.\\
			TSP & Traveling Salesman Problem.\\
			UAV & Unmanned Aerial Vehicles.\\
			UMSF & UAV Flight Mission Scheduling Framework.\\

		\end{tabular}
	\end{center}
\end{table}

\section{Introduction}
%Applications
Unmanned aerial vehicles (UAVs) have wide applications in recent years.
For example, UAVs are widely applied in disaster relief,
agriculture and cargo transportation, {and so on}.
The number of UAV flight activities has increased significantly worldwide.
%Numbers
According to Federal Aviation Administration (FAA),
1.7 million UAVs have been registered up to November 2020 \cite{M_introduction1}.
Besides, according to China Civil Aviation Working Conference in 2020,
there were more than 392,000 registered UAVs and 1.25 million
hours of commercial UAV flight activities in 2019 \cite{M_introduction2}.
It is also shown that the global UAV market size has reached to
9 billion dollars in 2019.
%Safety problems should be taken into consideration
With the massive applications of UAVs,
the anti-collision technologies of UAVs have attracted much attention
aiming to avoid the collisions between UAVs and other objects.

%Second graph
The anti-collision technologies of UAVs
are of vital importance to maintain the survivability of UAVs.
%Three parts of collision avoidance
The anti-collision technologies of UAVs generally consist of three procedures, namely,
obstacle sensing, collision prediction, and collision avoidance.
By sensing the surrounding environment,
UAV obtains the identity and location information of obstacles,
which can be used for collision prediction.
During obstacle sensing, the sensors in UAV obtain environmental information such as position and speed of obstacles. 
For UAV swarm, environmental information can be shared among UAVs via the wireless network connecting them. 
In the procedure of collision prediction,
UAV will determine whether it will collide with the obstacles.
If the UAV predicts that the collision will occur,
in the procedure of collision avoidance,
UAV will schedule the flight path to avoid the collision quickly.
%Challenges of collision avoidance
The anti-collision technologies of UAVs face several challenges.

{Firstly, UAVs fly in the three-dimensional (3-D) space. The complex trajectory of UAVs and the complex positional relationship between obstacles and UAVs bring great difficulty to the design of anti-collision algorithms. 
Secondly, UAVs will face more kinds of obstacles compared with ground vehicles, including not only the objects {on ground} such as trees, mountains, and buildings, but also flying objects such as birds and aircraft. Finally, most of the obstacles faced by UAVs are dynamic. The obstacle avoidance system needs to respond quickly according to the current situation and predict the trajectories of obstacles. 

{The existing anti-collision technologies are mostly for ground vehicles \cite{ground1,ground2, ground3}.}
Compared with UAVs, the anti-collision techniques for ground vehicles have the following features. 
Firstly, {ground} vehicles move in the two-dimensional (2-D) space, which are usually in the same plane as obstacles. 
Secondly, {the} obstacles faced by ground vehicles are mainly trees, cars, and buildings.
The kinds of obstacles of ground vehicles are smaller than that of UAVs. Finally, for ground vehicles, the prediction of collision is relatively simple. 
The existing anti-collision technologies for ground vehicles are difficult to satisfy the requirements of the anti-collision system for UAVs. {Nevertheless}, some of the anti-collision technologies for ground vehicles such as obstacle sensing methods can be enhanced and applied to UAVs.}
Academia and industry have paid much attention on
the anti-collision technologies of UAVs.
In this article, we summarize the existing anti-collision technologies
to provide guidelines for related researches.

%Existing surveys and our advantage
Regarding the research on anti-collision technologies of UAVs,
there are several survey articles \cite{M_introduction2.5, M_introduction3, M_introduction4, M_introduction6, L_introduction1, M_introduction7}.
Aswini \emph{et al.} in \cite{M_introduction2.5} reviewed UAV obstacle sensing methods applied in obstacle avoidance.
Ryan \emph{et al.} in \cite{M_introduction3} provided an overview of the cooperative UAV control,
i.e., collision avoidance of UAV swarm.
Jenie \emph{et al.} in \cite{M_introduction4} proposed a classification of
collision detection and avoidance approaches
for UAVs in an integrated airspace.
Chand \emph{et al.}
in \cite{M_introduction6} briefly introduced key technologies of sensing and collision avoidance for UAV.
%加内容 Unmanned Aerial Vehicles (UAVs): Collision Avoidance Systems and Approaches 在sensing 和 action 方面对防碰撞进行了调研
Shakhatreh \emph{et al.} in \cite{L_introduction1} introduced applications
and challenges of UAVs
in the civil field,
and studied collision avoidance methods of UAVs.
Yasin \emph{et al.} in \cite{M_introduction7} provided a survey of collision avoidance systems and approaches on sensing and collision avoidance techniques for UAVs.
However, the existing literature has some limitations.
None of the existing surveys have
reviewed the comprehensive anti-collision technologies of UAVs
regarding procedures of sensing, prediction,
and collision avoidance methods.
Besides, there are still no surveys
discussing laws, regulations, and
applications related to anti-collision technologies of UAVs.
Finally, the future trends of anti-collision technologies of UAVs are not analyzed in details.
All these issues will be addressed in this article, which aims to provide a comprehensive survey on anti-collision technologies of UAVs and provide the guidelines for the future trends.
The contributions of this article are as follows. 

\begin{itemize}
	
	\item The anti-collision technologies are classified into three procedures, i.e., obstacle sensing, collision prediction, and collision avoidance, which are reviewed respectively in details. The related laws and regulations to prevent anti-collision of UAVs are introduced.
	
	\item The future trends of anti-collision technology of UAV, such as fast obstacle sensing and fast wireless networking are analyzed. We discover that joint sensing and communication (JSC), namely, integrated sensing and communication (ISAC) technique is promising to improve the speed of sensing and networking. Thus JSC technique is {promising} to improve the anti-collision capability of UAVs.

\end{itemize}

\begin{figure}[!t]
	\label{Fig_Total}
	\centering
	\includegraphics[width=0.55\textwidth]{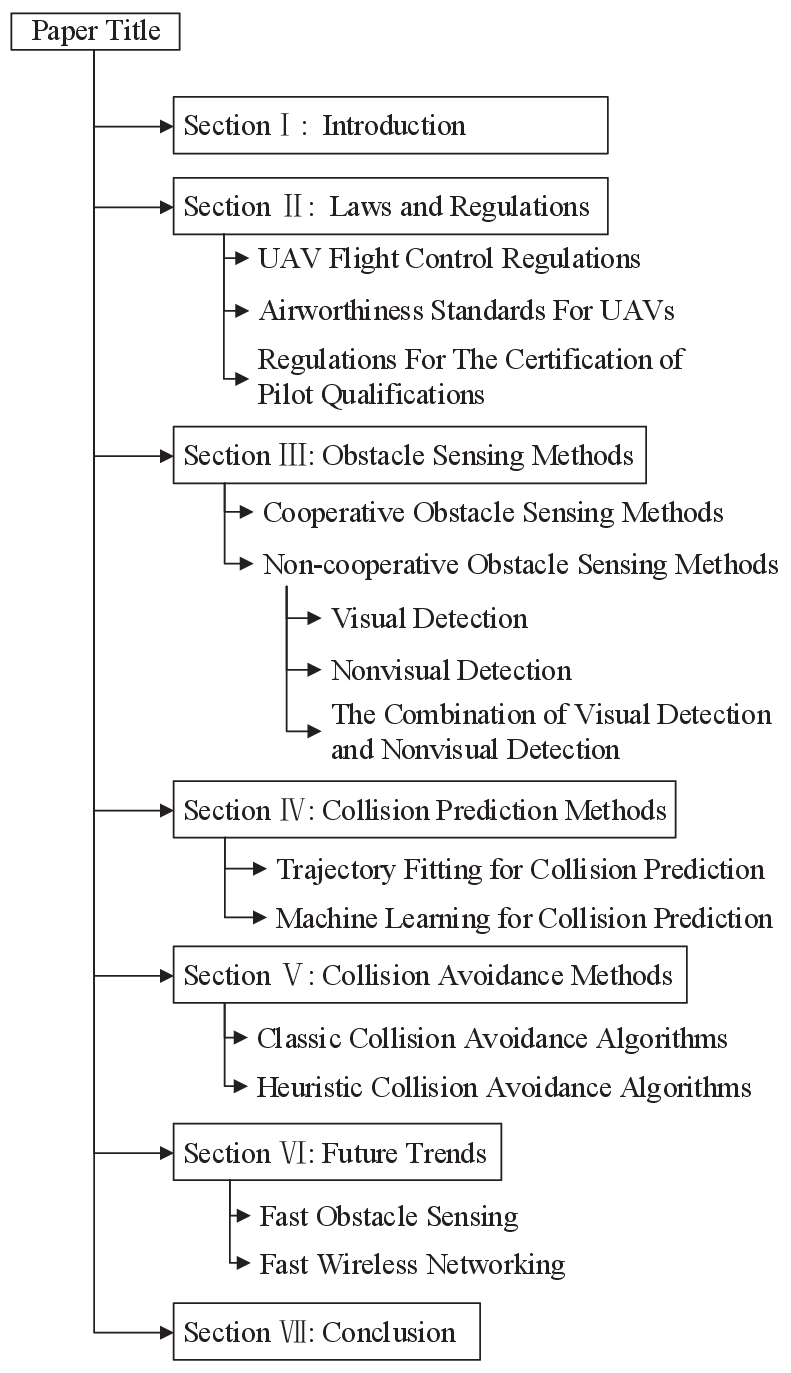}
	\caption{The organization of this paper}
	\label{The organization of this paper}
\end{figure}

%Structure
As shown in Fig. \ref{The organization of this paper},
the rest of this article is organized as follows.
In Section \uppercase\expandafter{\romannumeral2},
the laws and regulations related with aviation safety of UAVs are introduced.
In Section \uppercase\expandafter{\romannumeral3} to Section \uppercase\expandafter{\romannumeral5},
the anti-collision technologies of UAVs are reviewed.
Specifically, in Section \uppercase\expandafter{\romannumeral3},
the obstacle sensing methods are introduced,
which consists of cooperative and non-cooperative obstacle sensing methods.
Then, {the} collision prediction methods are reviewed in
Section \uppercase\expandafter{\romannumeral4}.
In Section \uppercase\expandafter{\romannumeral5},
the collision avoidance algorithms are reviewed.
In Section \uppercase\expandafter{\romannumeral6}, the future trends on UAV anti-collision technologies are
presented.
Finally, in Section \uppercase\expandafter{\romannumeral7}, we summarize this article.

%\begin{figure*}[!t]
%	\label{Fig_Total}
%	\centering
%	\includegraphics[width=0.8\textwidth]{BigPic.eps}
%	\caption{Structure of the Survey}
%	\label{Structure of the Survey}
%\end{figure*}

\section{Laws and Regulations}

Due to the severe environment and unstable control signal,
UAVs have a large probability to collide with obstacles.
UAV aviation safety laws and regulations are thus established to
guarantee the safety of UAVs and objects that might be hit by UAVs.
The UAV aviation safety laws and regulations can be divided into 
three categories, namely,
UAV flight control regulations,
UAV airworthiness standards, and
pilot qualification certification regulations, where
UAV flight control regulations regulate the flight activities of UAVs,
airworthiness standards provide
standards for the design and production of UAVs, and
pilot qualification certificate regulations imposes the requirements for UAV operators,
ensuring the flight safety.
The summary of laws and regulations of UAV is shown in Table \uppercase\expandafter{\romannumeral1}.

\begin{table*}[!t]
	\label{Table_1}
	\caption{Summary of laws and regulations}
	\renewcommand{\arraystretch}{1.1} %looser
	\begin{center}
		
		\begin{tabular}{|m{0.27\textwidth}<{\centering}|m{0.1\textwidth}<{\centering}|m{0.5\textwidth}|}
			\hline
			Category & Reference & \makecell[c]{One-sentence summary}\\

			\hline
			
			\multirow{4}[8]{\linewidth}{\makecell[c]{UAV Flight Control \\Regulations}}
			& \cite{L_laws0} & Interim Regulations on Flight Management of Unmanned Aircraft (Draft for Comment). \\
			\cline{2-3}
			& \cite{L_laws0_add} & Administrative Measures for Air Traffic Management of Civilian UAV Systems.\\
			\cline{2-3}
			&\cite{L_laws12} & Regulation on Dynamic Data Management of
			Light and Small Civil UAV flight.\\
			\cline{2-3}
			& \cite{L_laws7} & Draft of UAV System Remote Identification Rules.\\
			\hline
		
			\multirow{5}[13]{\linewidth}{\makecell[c]{Airworthiness Standards\\ For UAVs}}
			& \cite{L_laws8} &Airworthiness Standards for Medium and High-risk Unmanned Helicopter Systems (Trial).\\
			\cline{2-3}
			& \cite{L_laws9} &Airworthiness Standards for High-risk Fixed-wing Cargo UAV Systems (Trial).\\
			\cline{2-3}
			& \cite{L_laws10} &Large Cargo UAV airworthiness Standards.\\
			\cline{2-3}
			& \cite{L_laws11} &Guidance on UAV Airworthiness Certification Based on
			Operational Risks.\\
			\cline{2-3}
			& \cite{L_laws6} &Administrative Procedures for Airworthiness Certification of Civil Unmanned Aerial Vehicle Systems.\\
			\hline
			
			\multirow{2}[5]{\linewidth}{\makecell[c]{Regulations For The \\Certification of\\ Pilot Qualifications}}
			& \cite{L_laws1} &Regulations on the Administration of civilian UAV pilots.\\
			\cline{2-3}
			& \cite{L_laws3} &Airman Certificate Testing Service (ACTS) Award Notice.\\
			
			\hline
		\end{tabular}
	\end{center}
\end{table*}

\subsection{UAV Flight Control Regulations}

In order to regulate the flight activities of UAVs,
UAV flight control regulations have been established to regulate
the flight area, flight speed, and flight duration of UAVs.
It reduces the probability of path conflict between UAVs, 
forbids the flight behavior of UAVs in dangerous areas, 
thus fundamentally reduces the probability of collision.

For the management of the flight area,
the Civil Aviation Administration of China (CAAC) plans the flight airspace of UAVs in \cite{L_laws0},
and stipulates the horizontal and vertical ranges of the flight airspace based on
the characteristics of small and light UAVs.
Isolated airspace is also established to isolate UAV flight activities
from planes.
However, for medium or large UAVs that have passed safety certification,
light UAVs not exceeding the safe altitude,
and small UAVs with reliable monitoring and maintenance capabilities,
isolated airspace will not be reserved.
Regulation 107 in the ``Federal Aviation Regulations",
which was issued by FAA,
stipulates the flight range of civil small UAVs in the United States of America (USA),
and proposes that the flight activities of small UAVs must be kept within the sight of the operator.
In addition, the United States National Airspace System (NAS) divides airspace into
Class A, Class B, Class C, Class D, Class E, Class G, and some special airspace, in which
tower license is required for operations in airspace Class B, C, D, and E, but not in Class G, as shown in Table \uppercase\expandafter{\romannumeral2}.

For countries of the European Union (EU), international flight of cargo UAVs in airspace Class A, B, and C should take off and land at aerodromes under European Aviation Safety Agency (EASA)’s scope \cite{EULaws1}. In the United Kingdom (UK), Civil Aviation Authority (CAA) stipulates that the airspace division of airspace Class A-E is not applicable to the UAVs flying at low altitude. However, prohibited areas, restricted areas, dangerous areas and protected areas near the airport are forbidden to be entered by UAVs \cite{EULaws2}.

\begin{table*}[!t]
	\label{Table_1.5}
	\caption{Definition of the airspace classes\cite{L_laws_classes}}
	\renewcommand{\arraystretch}{1.1} %looser
	\begin{center}
		
		\begin{tabular}{|m{0.22\textwidth}<{\centering}|m{0.72\textwidth}|}
			\hline
			Classes & \makecell[c]{Definition} \\	\hline
			Class A & Class A includes the en route and high altitude environment of aircraft from one area to another in the identical country.\\
			\hline
			Class B & Class B airspace is defined at 29 high-density airports in the United States, aiming to manage the air traffic activities around the airfield. \\
			\hline
			Class C& Class C airspace is defined around the airport with airport traffic control tower and radar approach control.\\
			\hline
			Class D& Class D airspace is under the jurisdiction of a local Air Traffic Control Tower (ATCT).\\
			\hline
			Class E& Class E completely separates aircraft from other aircraft, aiming to provide air traffic services.\\
			\hline
			Class G& Class G airspace is defined as airspace not designated as class A, class B, class C, class D or class E.\\
			\hline

		\end{tabular}
	\end{center}
\end{table*}

Regarding the supervision of UAVs,
\cite{L_laws0_add}
promulgated by CAAC in 2016 puts forward requirements on the flight space and flight conditions of UAVs to prevent them from affecting civil aviation. 
The ``Regulation on Dynamic Data Management of Light and Small Civil UAV Flight" promulgated in 2019 points out that light UAVs, small UAVs, and plant protection UAVs in the airspace should be  monitored.
The regulation requires them to transmit dynamic data including flight code, duration of flight, position, speed, trajectory, {and so on}., 
in a certain format in real time for UAV management and research \cite{L_laws12}. 

FAA issued ``Draft of UAV System Remote Identification Rules" in December 2019\cite{L_laws7}. 
The rules stipulate that UAVs need to provide a third-party system with the order number, position, and other information during the flight, so as to ensure the flight safety. CAA stipulates that UAVs in the UK must have remote ID system, which can broadcast status and identity data including the verification code provided by CAA, the unique serial number of UAV, geographical location, flight route, and so on \cite{EULaws3}.

\subsection{Airworthiness Standards For UAVs}
The airworthiness standard of UAV specifies the performance requirements of UAV flight system, 
which prevents the UAV from collision due to the substandard performance.
CAAC has issued various types of UAV airworthiness standards in 
``The Airworthiness Standard of Medium and High Risk Unmanned Helicopter System (Trial)"\cite{L_laws8}, 
``The Airworthiness Standard of High Risk Cargo Fixed Wing UAVs (Trial)"\cite{L_laws9}
and ``The Airworthiness Standard of Large Cargo UAVs"\cite{L_laws10}. 
They stipulate the relevant UAV system performance requirements, including take-off, climb and glide performance, maneuverability, and so on. 
In addition, CAAC promulgated the ``Guidance on UAV Airworthiness Certification Based on Operational Risks" \cite{L_laws11}
, which puts forward different airworthiness management modes according to the risk level of UAV operation scenarios. 
The ``Risk Assessment Guide for Civil UAV System Airworthiness Certification Project (Draft for Comments)" issued in 2020 
proposes the authorization principles of risk system and product risk assessment. 
The product risk is based on the energy level of civil UAVs and the collision possibility level in the operating environment to form the risk matrix, which is applied to reveal the risk level of civil UAVs in different scenarios\cite{L_laws6}.

FAA does not require small UAVs to meet current airworthiness standards or obtain aircraft certification.
To ensure the normal operation of the safety system and the communication link between control station and UAV, 
the operator needs to carry out relevant inspections before flight, 
but the UAV must be registered. CAA suggests that large UAVs must have a valid certificate of airworthiness and small UAVs only need to have a permit to fly. Meanwhile, it is necessary to conduct safety assessment on UAV to ensure the reliability of UAV mission. Relevant laws are still being formulated \cite{EULaws3}.

\subsection{Regulations for The Certification of Pilot Qualifications}
In order to ensure the flight safety of UAVs, some laws and regulations stipulate that the UAV operators must acquire a license. 
CAAC passed the ``Regulations on the Management of Civil UAV Pilots" in 2019. 
For the civil UAVs with weight of more than 7kg, 
the pilot must obtain the driving license when satisfying the relevant standards \cite{L_laws1}. 
FAA has also regulated that if a pilot wants to operate a small UAV, the pilot needs to have a license or be directly supervised by a person with license, 
and the pilot must be at least 16 years old to acquire a pilot license \cite{L_laws3}.
EASA stipulates that for recreational or low-risk UAV activities, the operator only needs to comply with safe operation requirements. For civil UAV activities with high risks, operators need to obtain an operational authorization from the national competent authority before starting the operation. For future high-risk activities such as carrying passengers by UAV, the safety certification of the UAV operator and its UAV, as well as the licensing of the remote pilots are required to ensure safety \cite{EULaws4}.

In the anti-collision techniques for UAVs, the sensing for obstacles, the collision prediction and collision avoidance methods are essential. 
In the following sections, the three techniques are introduced respectively.
% \vbox{}

\section{Obstacle Sensing Methods}

Obstacle sensing is the first step of the anti-collision process for UAVs, 
through which UAV obtains the awareness of the surrounding environment, 
estimates the locations of obstacles, 
and provides prior information for anti-collision techniques of UAVs.

{There are two kinds of obstacles. The first kind of obstacles are environmental objects on the ground, such as trees, buildings, and mountains, which are usually static. The second kind of obstacles are flying objects such as birds and other aircraft, which are usually dynamic. Due to the complexity of the UAV trajectory, there is a high risk of collision between multiple UAVs if there are no anti-collision systems equipped on UAVs \cite{M_introduction6}.}

{There are cooperative and non-cooperative obstacle sensing methods. For cooperative sensing methods, UAVs share their state information or the information of surrounding obstacles. For non-cooperative sensing methods, UAVs relies on their own sensors to detect the obstacles without the participation of the communication process.}

\subsection{Cooperative Obstacle Sensing Methods}
%%-------------
Cooperative obstacle sensing system has been widely used in aircraft, and is gradually applied to UAV in recent years. For example, the Matrice 200 series and Mavic 2 Enterprise UAVs released by DJI have installed ADS-B system and can detect the ADS-B signals of aircraft miles away, then warns the aircraft if there is a premonition of collision. In this section, we introduce and classify the existing cooperative obstacle sensing methods and discuss the main challenges of them when applying to UAVs. There are mainly three types of cooperative anti-collision systems: Traffic Collision Avoidance System (TCAS), Automatic Dependent Surveillance – Broadcast system (ADS-B) and Flight Alarm (FLARM). TCAS and ADS-B are generally used for large and medium UAVs, while FLARM is used for small UAVs. The summary of cooperative obstacle sensing methods is shown in Table \uppercase\expandafter{\romannumeral3}.

\begin{table*}[!t]
	\label{Table_2}
	\caption{Summary of cooperative obstacle sensing methods}
	\renewcommand{\arraystretch}{1.3} %looser
	\begin{center}
		
		\begin{tabular}{|m{0.16\textwidth}<{\centering}|m{0.1\textwidth}<{\centering}|m{0.67\textwidth}|}
			\hline
			Methods & Reference &  \makecell[c]{One-sentence summary}\\
			
			\hline
			TCAS & \cite{H1-1} & Traffic Collision Avoidance System, a set of anti-collision systems installed in medium and large aircraft.\\
			\hline
			 ADS-B & \cite{H1-2} & Automatic Dependent Surveillance – Broadcast, a system transforming segments of aviation, which can effectively improve the cooperation between {aircraft} and
			enhance the performance of TCAS.\\
			\hline
			 FLARM & \cite{H1-12} & Flight Alarm, a traffic awareness and anti-collision technology 
			 for small UAVs with human operators on the ground.\\
			\hline
			
		\end{tabular}
	\end{center}
\end{table*}

TCAS is a set of anti-collision system installed in medium and large aircraft. At present, TCAS has become the standard equipment on the newly produced medium and large aircraft. The function of this system is to send inquiry signals to neighboring aircraft, and obtain the altitude, heading, and other data of the invading aircraft through the response of the onboard transponder system of the invading aircraft \cite{H1-1}. Through data analysis, the TCAS system determines the threat level of the invading aircraft. If there is a potential threat, the TCAS system will issue advisory prompts or vertical maneuver instructions to the pilot, which guide the pilot to avoid conflict with the invading aircraft.

ADS-B technology is widely applied in civil aviation. The shared information in ADS-B is the position information of aircraft and the information of conflict warning, track angle, route inflection point, {and so on}, as well as aircraft classification and identification information \cite{H1-2}. Compared with TCAS, the location report of ADS-B is spontaneously broadcast. Thus, the position report of the approaching UAV can be received and processed without inquiries between UAVs.

FLARM is a traffic awareness and anti-collision technology 
for small UAVs with human operators on the ground. The FLARM system obtains its own real-time position and altitude information through a built-in high-sensitivity GPS receiver and altimeter. 
Combined with speed and position information, FLARM can calculate an accurate predicted flight path. 
The path information is then broadcast to nearby {aircraft}. 
Meanwhile, the FLARM system receives the flight path information sent by the FLARM systems of nearby aircraft. 
If a collision is predicted, the FLARM system will send the warning information, as well as the direction and altitude difference information of the invading aircraft, to the connected FLARM information display of operator's device. 
When receiving the warning information, the operators can take corresponding actions to avoid possible collisions \cite{H1-12}. 

The existing cooperative anti-collision system can accurately detect the distance, altitude, and other information of the aircraft, and the detection distance is long. 
However, it can only detect aircraft installed with the same anti-collision system, and cannot detect obstacles such as mountains or trees in the environment.

\subsection{Non-cooperative Obstacle Sensing Methods}
The non-cooperative sensing system obtains the
location information of obstacles in the surrounding environment through sensors.
According to the types of sensors, the non-cooperative obstacle sensing methods are classified into visual detection, non-visual detection, and the combination of them. 
%-------还是有点长--------
The visual detection system includes monocular vision system and stereo vision system.  
The non-visual detection system
consists of the systems using active sensors including radar, laser, and the systems using passive sensors including thermal imaging, optoelectronics, and infrared. 
%---------------------
The fusion of visual and non-visual detection methods is also a promising approach to improve the accuracy and widen the scenarios of obstacle sensing.
The summary of non-cooperative obstacle sensing methods is shown in Table \uppercase\expandafter{\romannumeral4}.

\begin{table*}[!t]
	\label{Table_3}
	\caption{Summary of methods of non-cooperative obstacle sensing methods}
	\renewcommand{\arraystretch}{1.2} %looser
	\begin{center}
		
		\begin{tabular}{|m{0.16\textwidth}<{\centering}|m{0.16\textwidth}<{\centering}|m{0.1\textwidth}<{\centering}|m{0.48\textwidth}|}
			\hline
			Category & Methods & Reference & \makecell[c]{One-sentence summary}\\

			\hline
			\multirow{6}[29]{\linewidth}{\makecell[c]{Visual detection}}
			
			& \multirow{3}[8]{\linewidth}{\makecell[c]{Monocular \\vision}}
			& \cite{M_S_UC_13} & Research progress of obstacle detection based on monocular vision.\\
			\cline{3-4}
			&& \cite{M_S_UC_1} & Proposed a special attention mechanism to distinguish salient obstacles.\\
			\cline{3-4}
			&& \cite{M_S_UC_5} &Applied Gunnar-Farneback method to estimate the optical flow between two consecutive image frames.\\
			\cline{2-4}

			& \multirow{3}[6]{\linewidth}{\makecell[c]{Binocular stereo\\ vision}}
			&\cite{M_S_UC_20} &Designed a system based on binocular vision sensors for the detection and obstacle avoidance of micro-UAV in indoor environment.\\
			\cline{3-4}
			&& \cite{M_S_UC_3} & Applied binocular stereo vision combined with IFDS path planning.\\
			\cline{3-4}
			& & \cite{M_S_UC_7} & Applied six 4K cameras to build 3-D maps.\\
			\hline

			\multirow{5}[20]{\linewidth}{\makecell[c]{{Non-visual}\\ detection}}
			& \multirow{3}[8]{\linewidth}{\makecell[c]{Active type}}
			& \cite{M_S_UC_9} & Proposed a method to detect objects using ultrasonic sensors onboard.\\
			\cline{3-4}
			&&\cite{M_S_UC_16} &Proposed a low cost radar solution based on the coherent MIMO principles. \\
			\cline{3-4}
			&&\cite{M_S_UC_17} &Proposed a small-sized radar design according to the low-altitude high-speed environment of UAVs.\\
			\cline{2-4}
			
			&\multirow{2}[2]{\linewidth}{\makecell[c]{Passive type}} 
			&\cite{M_S_UC_18} &Proposed a system using IR and ultrasonic sensors which is capable of avoiding obstacles like walls and people.\\
			\cline{3-4}
			&&\cite{M_S_UC_18.5}&Applied LWIR cameras combined with NIR laser detection.\\
			\hline

			\multirow{3}[5]{\linewidth}{\makecell[c]{Combination}}
			& \multirow{3}[5]{\linewidth}{\makecell[c]{Combination}}
			& \cite{M_S_UC_10} &Applied monocular camera to estimate speed and laser radar to help avoid obstacles.\\
			\cline{3-4}
			&&\cite{M_S_UC_11} &Applied infrared light source to assist vision system.\\
			\cline{3-4}
			&&\cite{M_S_UC_12} &Combined visual cameras with ultrasonic sensors.\\
			\hline
		\end{tabular}
	\end{center}
\end{table*}

\subsubsection{Visual Detection}
%-----这里再加一段话描述视觉的特点---------%

There are generally two kinds of visual detection system, namely, monocular vision system and stereo vision system.

Monocular vision system employs a single camera to collect the obstacle information, 
which consists of three kinds of methods according to the application scenario. 
%%Three methods of monocular vision
%% 加参考文献

The first kind of methods can only detect specific targets, 
which use the prior knowledge of targets to recognize obstacles from the background image.
When the number of obstacles is huge or there are obstacles that are not consistent with the prior information, 
the accuracy of detection is poor. 
Therefore this kind of method is not widely used \cite{M_S_UC_13}.

The second kind of methods are inspired by the automatic focusing technology, 
which detects obstacles by focus/defocus or by changing the focal length.
For example, \cite{M_S_UC_1} proposed an obstacle sensing method inspired by human vision system (HVS).  
\cite{M_S_UC_1} adopted discrete cosine transform to recognize areas of prominent objects, 
so that UAVs can detect the obstacles in sight. 
This kind of method can simultaneously recognize and localize the obstacles. 
However, it will make mistakes when multiple obstacles co-exist, and the zooming ability of cameras should be high.
%%%

The third kind of methods detect obstacles by image difference or optical flow caused by the mobility of targets, which can only detect moving targets.
This kind of methods are also called motion parallax methods. 
Motion parallax refers to the phenomenon that the moving velocity of the target seems to be increasing with the decrease of the distance between observer and the target. 
\begin{figure}[!t]
	\centering
	\includegraphics[width=0.5\textwidth]{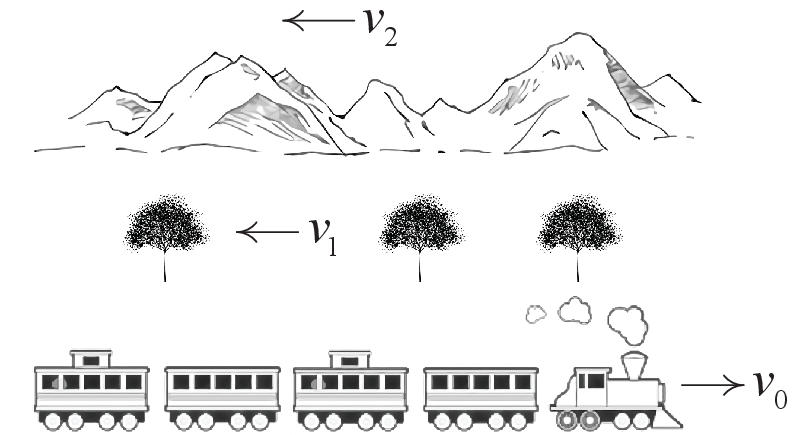}
	\caption{The velocity of nearby trees $v_1$ seems to be larger than that of distant mountains $v_2$ in the scenery outside the windows}
	\label{scenery outside the windows}
\end{figure}
As illustrated in Fig. \ref{scenery outside the windows}, when taking train, the trees near the window seem to move faster than the mountains far away. 
One of the motion parallax methods is to take advantage of the optical flow \cite{M_S_UC_OF}. 
\cite{M_S_UC_5} applied Gunnar-Farneback method to estimate the optical flow between two consecutive image frames. 
If the magnitude of the optical flow vector is smaller than a threshold, there are no obstacles in front of UAV.
Meanwhile, utilizing the state of UAV's quadrotor, the misjudgment caused by camera movement can be eliminated and the detection accuracy is improved.
However, the disadvantage of motion parallax methods is that the
processing time is long because multiple images of different times need to be processed.

\begin{figure}[!t]
	\centering
	\includegraphics[width=0.5\textwidth]{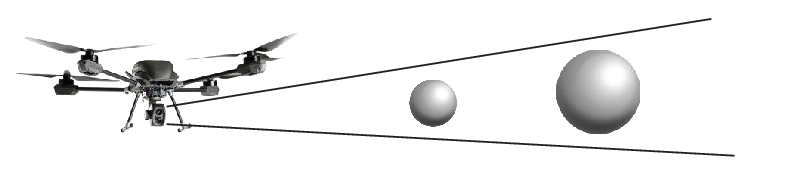}
	\caption{The disadvantage of the monocular vision}
	\label{The disadvantage of the monocular vision}
\end{figure}

Generally, the accuracy of monocular vision method is low, which is difficult to be overcome by algorithms.
%
%This is because that monocular vision cannot easily measure the true scale of the object. 
As illustrated in Fig. \ref{The disadvantage of the monocular vision}, the volume and distance of the  objects with the same area in two-dimensional (2-D) image of monocular vision system may be different. 
Thus, monocular vision systems are difficult to provide accurate information about distance and volume of the obstacle for collision avoidance.
Therefore, there are more and more literatures on stereo vision system. 
The stereo vision system applies multiple cameras to get pictures from different angles, where the system with two cameras is called binocular vision system. 
With the assistance of the pictures from two cameras, the size and location of obstacles can be estimated.
\cite{M_S_UC_20} designed a binocular vision system in a small UAV.
With distortion correction and binocular calibration of the pictures from two cameras, 
block matching algorithm is applied to obtain the disparity map, which reveals the 3-D depth information of target.
The target was then identified from the disparity map.

In \cite{M_S_UC_3}, the UAV with binocular vision system was applied in the power lines inspection scenario. 
They calculated the parallax value matrix through left and right images from two cameras 
and used the matrix along with camera parameters to obtain relative 3-D coordinates of targets.

Besides the binocular vision system, multiple cameras are installed in UAV to improve the accuracy of obstacle detection. 
Skydio UAV \cite{M_S_UC_7} installs six 4K cameras to construct  a 3-D map of surrounding environment including trees, people, buildings, {and so on}. 
Pictures from multiple angles enable UAVs to accurately obtain the location and distance of obstacles in all directions.

In addition to the advantage in accuracy of obstacle detection,
stereo vision system has a large field of vision because of the application of multiple cameras,  
which gather comprehensive environment information \cite{M_S_UC_14}. 
However, the complexity of hardware and algorithms of stereo vision system brings high cost and large power consumption, which is the main challenge for the application of stereo vision system in low-cost civilian UAVs.

Because of the mature algorithms and relatively low cost, visual detection systems are widely adopted by UAV manufacturers \cite{M_S_UC_15}. 
However, visual detection requires a good sight. 
Darkness, bad weather, or reflection from water surface like peaceful water will bring great challenges for the information acquisition in visual detection systems.

\subsubsection{{Non-visual} Detection}
{Non-visual} detection sensors consist of active sensors and passive sensors.

The active sensors, including radar, laser, and ultrasonic sensors, 
detect obstacles by transmitting signals to targets, receiving and analyzing echo signals. 
Laser has a large detect range, 
and can obtain not only distance, but also azimuth information. 
Besides, it has a great penetrating power when facing smoke. 
{UAVs of Walkera \cite{ResponseLetter_Walkera} apply laser for sensing, which realized a maximum sensing range of 40 m.}
However, laser is greatly affected by the bad weather.
On the contrary, ultrasonic sensors are not easy to be disturbed by smoke or gas. 
\cite{M_S_UC_9} proposed a method for UAVs to detect obstacles on the ground using ultrasonic sensors. 
The authors confirmed that it is feasible to detect wall by ultrasonic sensors, with error of about 40 mm. 

Radar provides the sensing information of ranging, direction, and closing speed\cite{M_S_UC_16}, and the sensing performance is stable regardless of the surrounding environment. 
Therefore, among the active sensors, radar is widely applied in UAVs for obstacle detection.
\cite{M_S_UC_16} designed a radar system based on multiple-input and multiple-output (MIMO), which achieves accurate angle estimation and large detection range. 
\cite{M_S_UC_17} proposed a small-sized radar design according to the low-altitude high-speed environment of UAVs, 
whose successful detection probability  is more than 90$\%$ on the condition that the maximum speed of UAV is 440 km/h and the minimum detection range is 2.8 km. 
In the industry, DJI proposed agriculture solution package containing new flight controller and radar sensing system \cite{M_S_UC_8}. 
The system equips high-precision radar which is able to detect obstacles behind or in front of the UAV at a range of 1.5 to 30 m. 
The radar cannot detect objects on the left or the right, 
because plant protection UAVs don't need to make a turn frequently.
{The agricultural UAVs of DJI \cite{M_S_UC_8} and XAG \cite{ResponseLetter_XAG} apply radar for obstacle sensing, which can sense obstacles in the range of 1.5 ~ 30 m, whose measurement accuracy is smaller than 10 cm.}

The passive sensors consist of infrared radiation (IR) sensors, thermal imaging cameras, electro optical sensors, {and so on}. 
Unlike active sensors, passive sensors don't transmit signals by themselves, but collecting signals emitted from the objects. 
For example, the infrared radiation sensors can detect targets by the emitted heat. 
The advantage of passive sensors is that they are usually much cheaper than active sensors. 
However, the signals gathered by passive sensors usually contain much noise \cite{M_S_UC_18}. 
\cite{M_S_UC_18} proposed a system using IR and ultrasonic sensors which is capable of avoiding obstacles like walls and people, 
and proved that cheap sensors can also achieve efficient detection. 
In fact, IR sensors perform even better compared with ultrasonic sensors because the ultrasonic sensors cannot detect human bodies reliably \cite{M_S_UC_18}. 
However, for practical applications, 
passive sensors are not usually applied alone in the anti-collision system. 
On the contrary, they cooperate with other sensors like radars to detect obstacles. 
For example, \cite{M_S_UC_18.5} applied a long wavelength infrared (LWIR) camera to achieve a fast detection of a thermal object, 
and more precise detection was implemented by the Near Infrared (NIR) laser detector.

To sum up, compared with visual detection, 
{non-visual} detection can also be effective of sensing when the field of vision is poor. 
However, they have lower accuracy than the visual detection. 
Therefore, plenty of UAVs nowadays choose to apply the combination of visual detection and {non-visual} detection.
\begin{table*}[!t]
	\label{Table_Sensing_PandC}
	\caption{Pros and cons of obstacle sensing methods}
	\renewcommand{\arraystretch}{1.1} %looser
	\begin{center}
		
		\begin{tabular}{|m{0.12\textwidth}|m{0.1\textwidth}|m{0.24\textwidth}|m{0.24\textwidth}|m{0.18\textwidth}|}
			\hline
			\multicolumn{2}{|c|}{Name of the methods}& \multicolumn{1}{c|}{Pros}& \multicolumn{1}{c|}{Cons}& \multicolumn{1}{c|}{Common features} \\
			\hline
			Cooperative Obstacle Sensing Methods &
			TCAS ADS-B FLARM &$\bullet$  High accuracy. & $\bullet$ Can only be used for UAVs equipped with the same system.& -\\
			\hline
			
			\multirow{3}[10]{\linewidth}{Visual Detection} 
			& Monocular Vision System  & $\bullet$ Low cost. &$\bullet$ Affected by darkness or strong light.& \vspace{0.5em}\multirow{2}[0]{\linewidth}{$\bullet$  The accuracy is related to the performance of the image processing algorithm.}\\
			\cline{2-4}
			&\multirow{2}[0]{\linewidth}{Stereo Vision System} & \multirow{2}[3]{\linewidth}{$\bullet$ Improved accuracy compared with monocular vision system.} &
			\vspace{1em}	$\bullet$  Affected by darkness or strong light. &\\
			&&&$\bullet$ High cost.\vspace{1em}&\\
			\hline
			
			\multirow{4}[15]{\linewidth}{{Non-visual} Detection} 
			&\multirow{3}[10]{\linewidth}{Active Sensors}  & \vspace{1em}$\bullet$ Radar: Great detection range. &\vspace{1em} $\bullet$ Radar: Sensitive to electromagnetic interference.& \multirow{4}[17]{\linewidth}{$\bullet$ The accuracy is related to the quality of sensors.} \\
			&&\vspace{0.5em}$\bullet$ Laser: High resolution. &$\bullet$ Laser: High cost.&\\
			&&$\bullet$ Ultrasonic Sensors: Low cost. & $\bullet$ Ultrasonic Sensors: Sensitive to the meteorologic environment.&\\
			\cline{2-4}
			&Passive Sensors & $\bullet$ Low cost. & $\bullet$ Low accuracy.&\\

			\hline	
		\end{tabular}
	\end{center}
\end{table*}
\subsubsection{The Combination of Visual Detection and {Non-visual} Detection}
The combination of visual and {non-visual} sensors could exploit both of their advantages. 
When designing the UAV indoor navigation system, \cite{M_S_UC_10} implemented a system with monocular camera and laser rangefinder radar. 
Experiment results show that the estimation error of location is about 0.5 m, and the error of speed estimation is 0.2 m/s. 
Companies like XAG apply binocular vision system to sense the environment information 
and apply infrared light sensors to detect obstacles at night \cite{M_S_UC_11}.  
{The \emph{Guidance} system of DJI \cite{M_S_UC_12} combines stereo vision system and ultrasonic sensors for sensing, which can detect obstacles with a speed of 0 $\sim$ 16 m/s at 0.2 $\sim$ 20 m. It has a speed measurement accuracy of 0.04 m/s and a positioning accuracy of 0.05 m.}

The pros and cons for collision avoidance methods are shown in Table \uppercase\expandafter{\romannumeral5}.

\section{Collision Prediction Methods}

Collision prediction is the second step of anti-collision process. 
UAV processes the information obtained from obstacle sensing, 
predicts whether there will be collision, 
and judges whether it needs to avoid collision according to the prediction results.
{The obstacles studied in collision prediction are mostly dynamic obstacles such as birds or other UAVs \cite{H_P_1, H_P_2,H_P_3,H_P_4, H_P_5, H_P_6,H_P_7, M_P_8}, because the collision probabilities of stationary obstacles are easy to estimate.}

{There are many ways for the classification of UAV collision prediction methods. According to the types of obstacles, the collision prediction methods can be classified into collision prediction between UAV and obstacles, and collision prediction between UAVs. According to the priori information required for prediction, the collision prediction methods can be classified into the trajectory prediction of obstacles, the trajectory prediction of UAVs, the prediction of collision probability, {and so on}. 

In this article, according to the prediction algorithms, the collision prediction methods are classified into trajectory fitting methods and machine learning methods. 

The {trajectory fitting} methods apply criteria such as the least-square criterion to fit trajectory points into a predetermined form of expression to predict the future trajectory \cite{H_P_1, H_P_2,H_P_3,H_P_4}. Since the form of the fitting function is fixed, such methods are difficult to predict precisely in complex environment of UAV. However, this method has low computational overhead and is suitable to low-cost UAVs.

The machine learning methods apply machine learning algorithms such as Recurrent Neural Network (RNN) and reinforcement learning to learn the previous motion information and output parameters such as future trajectory or collision probability \cite{H_P_5, H_P_6,H_P_7}. Generally speaking, the machine learning methods can better adapt to the complex flight environment of UAV since nonlinear problems can be better solved by machine learning. However, machine learning methods will increase the energy consumption on calculation and reduce the endurance of UAV.}

%\begin{figure}[!t]
%	\centering
%	\includegraphics[width=0.4\textwidth]{collisionPrediction.eps}
%	\caption{Types of collision prediction}
%	\label{Collision preiction}
%\end{figure}
\subsection{Trajectory Fitting for Collision Prediction}
The trajectory fitting can be applied to predict the flight path of UAV according to the kinematics and dynamic features of UAV.

\cite{H_P_1} proposed a UAV collision prediction method based on sliding window polynomial fitting algorithm. 
Firstly, the sliding window polynomial fitting method is applied to predict the obstacles' trajectory. 
Then, flight information of UAVs and obstacles is used to predict the collision. 
The algorithm needs a small number of historical data 
to construct the polynomial fitting equation, 
so as to realize the real-time collision prediction. 
This algorithm has been widely applied in the field of trajectory prediction \cite{H_P_2}.

\cite{H_P_3} proposed the Pythagorean-Hodograph (PH) and eight Bernstein-Bezier polynomials based method to predict the trajectory of UAV swarm for collision prediction. 
Each UAV established its trajectory based on the five supporting points of the leading UAV, predicted the trajectories of the UAV swarm, and avoided collision by fitting the trajectory curves of the UAV swarm. 
\cite{H_P_4} proposed a UAV path planning method based on an efficient quadratic Bezier curve. 
The algorithm used the starting point, end point and intermediate control points to form a quadratic Bezier curve to generate the trajectory prediction of the UAV, 
which can properly change the derived intermediate points to control the trajectory of UAV to avoid collisions with identified obstacles.

\subsection{Machine Learning for Collision Prediction}
UAVs estimate the movement of dynamic obstacles based on machine learning, so that UAVs can make collision prediction.

\cite{H_P_5} proposed a Long Short-Term Memory (LSTM) RNN based collision prediction method. 
This algorithm predicts the movement state of obstacles based on the most recent observations, which can greatly improve the prediction accuracy.

\cite{H_P_7} proposed a new model-based reinforcement learning algorithm TEXPLORE, which can predict the trajectory of UAVs in unknown or uncertain environments to reduce the probability of collisions.

Combining Deep Reinforcement Learning (DRL), \cite{H_P_6} proposed a DRL assisted method to train a deep Q network (DQN) to predict the trajectory of obstacles and thus fulfill collision prediction.

{The above collision prediction methods using machine learning are based on simulation environments, in which the dynamic obstacles are often assumed to be movable points, which is not appropriate for practical scenarios. Therefore, \cite{M_P_8} proposed a collision avoidance algorithm where neural network pipeline (NNP) is applied to predict the possible collision. Besides, an object motion estimator (OME) using optical flow is proposed to identify dynamic objects and estimate their trajectories in video streams. The algorithm has high accuracy (about 2\%) and fast processing speed.}

\section{Collision Avoidance Methods}

Collision avoidance is the core step of UAV anti-collision technology.
UAVs process the prior information obtained by obstacle sensing and collision prediction to guarantee the safe flight path without collision.
Generally speaking, collision avoidance problem is equivalent to path planning problem with obstacles in the flight path, 
because both of their purposes are to make UAV avoid obstacles with certain prior knowledge.
Generally speaking, collision avoidance methods can be classified into two categories: classic methods and heuristic methods.

\subsection{Classic Collision Avoidance Algorithms}

Classic collision avoidance algorithms of UAVs are developed from the robot obstacle avoidance algorithms, which is relatively mature.
This kind of algorithm consists of geometric methods, graph theory methods, artificial potential field methods, and methods based on path selection.
The summary of classic collision avoidance algorithms is shown in Table \uppercase\expandafter{\romannumeral6}.
\begin{table*}[!t]
	\label{Table_5}
	\caption{Summary of classic collision avoidance algorithms}
	\renewcommand{\arraystretch}{1.2} %looser
	\begin{center}
		
		\begin{tabular}{|m{0.17\textwidth}<{\centering}|m{0.17\textwidth}<{\centering}|m{0.1\textwidth}<{\centering}|m{0.4\textwidth}|}
			\hline
			Category & Methods & Reference & \makecell[c]{One-sentence summary} \\
			\hline
			
			\multirow{6}[15]{\linewidth}{\makecell[c]{\makecell[c]{Geometric\\ methods}}}
			&\multirow{3}[1]{\linewidth}{\makecell[c]{Collision cone}} &
			\cite{M_CA_CM_2} &Chose the direction to make a turn by the angle between the direction vector and the tangential lines of the obstacle.\\
			\cline{3-4}
			&&\cite{M_CA_CM_3}&Found the best aiming point of the collision cone.\\
			\cline{3-4}
			&&\cite{M_CollisionCone1} & Extended the collision cone to 3-D scenario.\\
			\cline{2-4}
			
			&\multirow{3}[8]{\linewidth}{\makecell[c]{Dubins path}}
			&\cite{M_CA_CM_4}&Applied the Dubins path algorithm to get all the paths and find the shortest one.\\
			\cline{3-4}
			&&\cite{M_CA_CM_5}&Applied the Dubins path algorithm on ground obstacle avoidance.\\
			\cline{3-4}
			&&\cite{M_CA_CM_5.5}&Extended the Dubins path algorithm on avoiding dynamic obstacles.\\
			\hline
			
			\multirow{3}[8]{\linewidth}{\makecell[c]{Graph theory\\ methods}}
			&\multirow{2}[5]{\linewidth}{\makecell[c]{Voronoi graph \\ search algorithm}}
			&\cite{M_CA_CM_6}&Presented the construction of the Voronoi graph search algorithm and estimated its performance.\\
			\cline{3-4}
			&&\cite{M_CA_CM_7}&An improvement and application of the Voronoi graph search algorithm.\\
			\cline{2-4}
			&\makecell[c]{Laguerre graph \\ search algorithm} &\cite{M_CA_CM_8}& Applied the Laguerre graph search algorithm to plan the flight path before a flight mission.\\
			\hline
			
			\multirow{4}[25]{\linewidth}{\makecell[c]{Artificial Potential \\ Field}}
			&\multirow{4}[25]{\linewidth}{\makecell[c]{Artificial Potential \\ Field}} & \cite{M_CA_CM_9} &Applied the potential field on the UAV collision avoidance and analyzed its performance\\
			\cline{3-4}
			&&\cite{M_CA_CM_10}&Improved the algorithm by changing the repulsive potential function according to the detection of the field.\\
			\cline{3-4}
			&&\cite{M_CA_CM_11}&Added weights on different obstacles and imported virtual targets in points of local minimum problems. \\
			\cline{3-4}
			&& \cite{M_add_APF1} & Introduced a destination switching scheme in the traditional artificial potential field method. \\
			\hline
			
			\multirow{2}[5]{\linewidth}{\makecell[c]{Path Selection \\ based Methods}}
			& \multirow{2}[5]{\linewidth}{\makecell[c]{Path Selection \\ based Methods}}
			& \cite{M_CA_CM_add1} & Proposed an improved RRT algorithm which shorten the time of path selection.\\
			\cline{3-4}
			&& \cite{M_CA_CM_add2} & Proposed an optimization based online collision avoidance algorithm for Internet of Drones (IOD) formation.\\

%			\hline
%			Dynamic fences & Dynamic fences&\cite{M_CA_CM_12}&Added fences on obstacles and no-fly zones.\\
			
			\hline
		\end{tabular}
	\end{center}
\end{table*}
\subsubsection{Geometric Methods}
Geometric method is an effective method to avoid collision, which takes advantage of relative kinematic geometric relationships between UAVs and obstacles.

(1) Collision {Cone}

\begin{figure}[!t]
	\centering
	\includegraphics[width=0.45\textwidth]{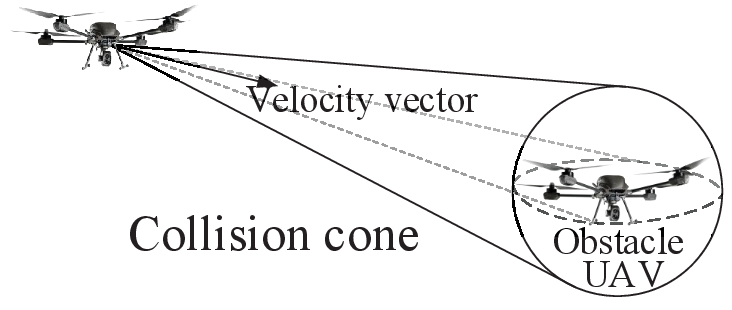}
	\caption{Collision cone}
	\label{Collision Cone}
\end{figure}

As illustrated in Fig. \ref{Collision Cone}, the collision cone is formed by a set of tangent lines from the UAV to the risk area, where the risk area is a sphere area around the obstacle.
If the velocity vector of the UAV is detected within the collision cone, the UAV will fly out of the collision cone as fast as possible to prevent collision according to the collision cone algorithm.

Specifically, {for stationary obstacles,} \cite{M_CA_CM_2} applied collision cone algorithm and established a collision cone when obstacles appear in the flight path of UAV. 
When a collision is about to occur, the destination of flight path is temporally set as the closest tangent point.
Then, the UAV will fly along the tangent line through this tangent point.
When the UAV reaches this point, the destination of the UAV will be reset to the original destination.
The UAV will not enter the collision cone again due to the inertia, so that it can bypass the obstacle.
 
{For dynamic obstacles,} \cite{M_CA_CM_3} constructed a 2-D collision cone and searches for the minimum angle $\theta_d$ between the {relative} velocity vector of the UAV and the hypotenuses of the cone. 
The {direction} of the UAV will be rotated immediately with an angle greater than $\theta_d$ to ensure that the UAV flies out of the collision cone. 
{\cite{M_CollisionCone1} extended the collision cone method to 3-D scenario. A pyramid collision cone is established to avoid cube obstacles. 
The flight test is taken indoors with static obstacles in the flight path, which shows that the algorithm can successfully avoid static obstacles in 3-D space.}

Collision cone is usually applied to deal with the obstacles nearby and can be applied in the scenarios without too many obstacles. {Besides, due to the low computational complexity and high real-time performance, the collision cone methods have good performance on collision avoidance of dynamic obstacles.}
%Collision cone method solves the collision problem by making small changes to the original path to update current positions.

(2) Dubins {Path}

\begin{figure}[!t]
	\centering
	\includegraphics[width=0.5\textwidth]{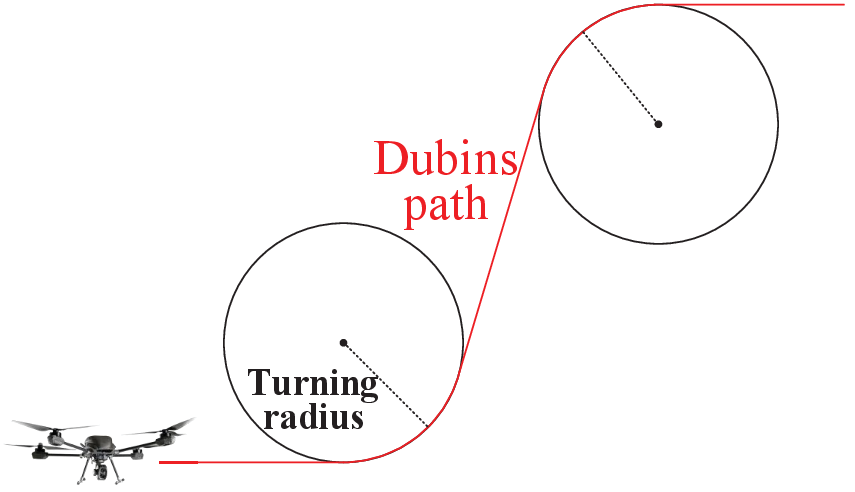}
	\caption{An example of Dubins path}
	\label{DubinsPath}
\end{figure}

The Dubins path is the shortest path between two positions considering the turning radius and the speed direction of vehicles, 
which was proposed by Dubins \cite{M_CA_CM_3.5}.
Because of the turning radius, the Dubins path is usually constructed by a series of arcs and the tangents between the arcs, as illustrated in Fig. \ref{DubinsPath}.
The Dubins path is widely used in collision avoidance algorithms of UAVs. 
Considering 2-D path planning problem of UAV, \cite{M_CA_CM_4} constructed Dubins paths between the vertices of the polygon envelopes of the obstacles in the environment.
By flying along the Dubins paths, UAV can reach the destination without collision.
The path planning problem without collision is thus transformed into the shortest path problem of an undirected graph, 
which can be solved using Dijkstra's algorithm.

Instead of concentrating on the targets in the air, 
\cite{M_CA_CM_5} applied Dubins path algorithm on ground obstacle avoidance
and carried out the software-in-the-loop experiment.
The results show that the flight path deviation of UAV at 10 m/s is lower than that at 15 m/s, 
because Dubins paths have a large number of arc turning paths and UAV is more difficult to control accurately when passing these paths at high speed.

{One of the disadvantages of the Dubins path algorithm is that it is difficult to be applied in the scenarios with dynamic obstacles.}
Therefore, \cite{M_CA_CM_5.5} applied a variation of Rapidly-exploring Random Tree (RRT) in the collision avoidance algorithm using 3-D Dubins path.
When the current path is predicted to be collided with the updated obstacle position, 
the algorithm will update the path according to the new obstacle position, so as to avoid dynamic obstacles.

\subsubsection{Graph Theory Methods}
The graph theory based collision avoidance methods consist of Voronoi graph search algorithm, Laguerre graph search algorithm, {and so on}. 
This kind of methods divide the space into a graph 
and traverse the graph to find the best path from source to destination.
{The graph theory methods are difficult to be applied to avoid collision with dynamic obstacles, because the global graph needs to be reconstructed after the change of the positions of obstacles, which brings huge time consumption.}

(1) Voronoi {Graph Search Algorithm}

\begin{figure}[!t]
	\centering
	\includegraphics[width=0.4\textwidth]{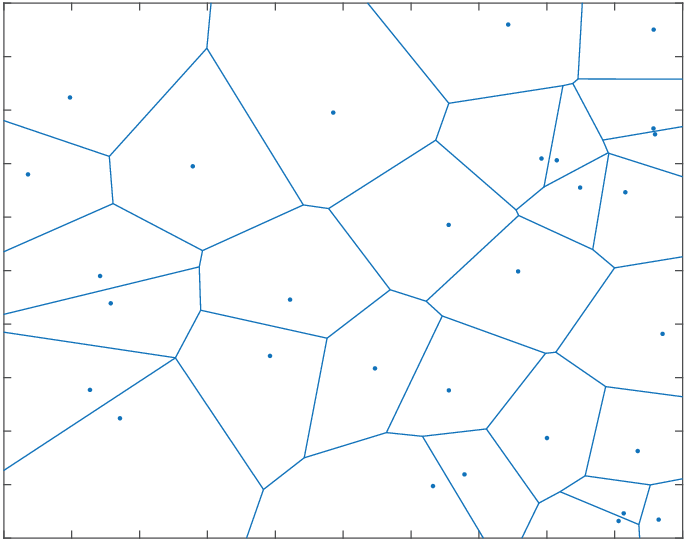}
	\caption{A Voronoi graph, where points represent obstacles and lines represent possible flight paths}
	\label{A Voronoi Graph}
\end{figure}

As illustrated in Fig. \ref{A Voronoi Graph}, each edge in the Voronoi graph is a vertical bisector of two points.
The Voronoi graph search algorithm for collision avoidance regards obstacles as points,
regards edges of the Voronoi graph as flight paths,
and applies algorithms such as Dijkstra algorithm to search for the shortest path from source to destination. 
The algorithm was widely used by robots and is applied in UAVs' path planning recently. 
\cite{M_CA_CM_6} designed the Voronoi graph search algorithm for UAVs and estimated its performance. 
It shows that the time consumption of the algorithm mainly comes from the establishment and smoothing of the Voronoi graph, and will increase sharply when the number of obstacles increases.

Furthermore, \cite{M_CA_CM_7} improved the Dijkstra algorithm after the Voronoi graph was constructed.
When the destination changes, the algorithm can update the flight path in time, which improves the applicability of the Voronoi graph search algorithm.

(2) Laguerre {Graph Search Algorithm}

The Laguerre graph is a variation of Voronoi graph. 
%Instead of constructed by polygons of vertical bisectors, 
%a Laguerre graph's line is decided by circular range of threat areas. 
%Each line crosses the interspaces of circular ranges and forms the polygons of the Laguerre graph. 
The distance from each edge to the adjacent points is no longer the same, but proportional to the weight of the adjacent points.
When applied for collision avoidance, the weight of the points is set by the radius of circumscribed circle of obstacles, 
which avoids the disadvantage of Voronoi graph that the flight path may cross the obstacles when the obstacles are too large or too close.

Similar to the Voronoi graph search algorithm, 
the Laguerre graph search algorithm first constructs the Laguerre graph, 
and then applies shortest path algorithms on the graph. 
For example, \cite{M_CA_CM_8} applied this algorithm in path planning before a flight mission. 
The irregular obstacles near the UAV is simplified as circles, which reduces the complexity of the algorithm. 
Meanwhile, a path optimization algorithm was proposed to make the flight path smooth.
\cite{M_CA_CM_8} showed that the Laguerre search graph algorithm applied in path planning is both fast and accurate. 
However, simplifying the obstacles into circles may make some of them intersect, which will leave no space for a potential flight path, so that a longer flight path will be generated.

\subsubsection{Artificial Potential Field}

Using the potential function, the artificial potential field algorithm establishes a repulsive force field for each obstacle.
With the guidance of force fields, the UAVs avoid the obstacles and reach the destinations.
%{The artificial potential field is difficult to be applied to collision avoidance of dynamic obstacles, because the algorithm needs to calculate the distribution of global potential field in advance to plan the path.}

\cite{M_CA_CM_9} applied the Gaussian mixture model to build the physical model of obstacles, 
and applied expectation-maximization (EM) algorithm to modify the potential fields by obstacles' information. 
%Then they applied the potential field on the UAV collision avoidance and analyzed its performance. 
By updating the Gaussian mixture model of the obstacles, the method proposed in \cite{M_CA_CM_9} can be used with dynamic obstacles. 
However, the potential field algorithm easily generates the local minimum solution in complex environment. 
%\cite{M_CA_CM_10} improved the algorithm by changing the repulsive potential function according to the detection of the field. 
Facing such dilemma, \cite{M_CA_CM_10} proposed a new repulsive potential function, which added an adjustment factor $m$, so that the optimal flight path can be obtained by adjusting $m$. 
\cite{M_CA_CM_11} added weights on different obstacles to represent different levels of threat. 
To overcome local minimum problem, 
\cite{M_CA_CM_11} placed virtual obstacles in the points with possible local minimum solutions. 
{Considering obstacle avoidance problem in a UAV formation, \cite{M_add_APF1} introduced a destination switching scheme in the traditional artificial potential field method.
The destinations of UAVs are switched when the distance between UAVs is too small, which avoids local optimization points caused by nearby UAVs.
The flight test is taken outdoors, which verifies the rationality of the proposed algorithm.}

\subsubsection{Path Selection based Methods}
{For path selection based methods, the paths are firstly randomly generated between the UAV and the destination.
The obstacle-free path is then selected by RRT, optimization, or other algorithms.
\cite{M_CA_CM_add1} proposed an improved RRT algorithm, which shortens the time of path selection of the traditional RRT algorithm.
Both stationary and dynamic obstacles are considered, where dynamic obstacles are represented by reachable sets, i.e. the areas that may be reached in a period of time.
The problem of avoiding dynamic obstacles is transformed into avoiding stationary reachable sets.
However, only obstacles moving with constant velocity can be represented by the reachable sets, so the proposed method lacks the ability to deal with emergencies.
\cite{M_CA_CM_add2} proposed an online collision avoidance algorithm for Internet of Drones (IOD) formation.
The total flight time is divided into small time slots.
In each time slot, multiple paths are generated randomly and the optimal path is selected by gradient optimization and various constrains.
The proposed algorithm can deal with dynamic obstacles with changeable velocity, because it only relies on instantaneous locations of obstacles in each time slot.
}

\subsection{Heuristic Collision Avoidance Algorithms}
Compared with the classical methods searching the optimal solution, 
heuristic algorithms aim to find the appropriate solution according to historical experience. 
The solution of heuristic algorithm is probably approximately optimal.
Heuristic collision avoidance algorithms consist of neural network algorithm, reinforcement learning, genetic algorithm, simulated annealing algorithm, colony algorithm, and so on.
The summary of heuristic collision avoidance algorithms is shown in Table \uppercase\expandafter{\romannumeral7} and \uppercase\expandafter{\romannumeral8}.
\begin{table*}[!t]
	\label{Table_4}
	\caption{Summary of heuristic collision avoidance algorithms(1)}
	\renewcommand{\arraystretch}{1.2} %looser
	\begin{center}
		
		\begin{tabular}{|m{0.16\textwidth}<{\centering}|m{0.1\textwidth}<{\centering}|m{0.65\textwidth}|}
			\hline
			Category  & Reference & \makecell[c]{One-sentence summary}\\
			\hline
			\multirow{6}[15]{\linewidth}{\makecell[c]{Neural network\\ algorithm}}
			&\cite{L_nonH_6}&Proposed a path planning algorithm based on ANN.\\
			\cline{2-3}
			&\cite{L_nonH_8}&Applied G-FCNN algorithm to evaluate the path.\\
			\cline{2-3}
			&\cite{L_nonH_1}&Applied neural network and Q-learning algorithm to solve the 3-D collision avoidance problem.\\
			\cline{2-3}
			&\cite{L_nonH_2}&Applied CGAN network to build the depth map used for further collision avoidance.\\
			\cline{2-3}
			&\cite{L_nonH_3}&The depth map was predicted by using CNN network.\\
			\cline{2-3}
			&\cite{L_nonH_4}&An algorithm combining the depth neural network and the epipolar geometry was proposed.\\
			\cline{2-3}
			&\cite{M_add_NN1} &Applied CNN to estimate the optical flow of the pictures.\\
			\hline

			\multirow{4}[14]{\linewidth}{\makecell[c]{Reinforcement \\ learning}}
			&\cite{L_nonH_7}&Introduced the ARE method on navigation and obstacle avoidance problems.\\
			\cline{2-3}
			&\cite{L_nonH_5}&Proposed Advantaged Actor - Critic algorithm innovatively.\\
			\cline{2-3}
			&\cite{L_nonH_26}&A multi-UAV anti-collision framework based on reinforcement learning was proposed.\\
			\cline{2-3}
			&\cite{L_nonH_27}&Proposed the two-stage reinforcement learning method.\\
			\cline{2-3}
			&\cite{M_add_27.5} & Proposed a distributed DRL framework on collision avoidance of dynamic obstacles.\\
			\hline

			\multirow{6}[12]{\linewidth}{\makecell[c]{Genetic\\ algorithm}}
			&\cite{L_nonH_10}&Applied Bellman Ford algorithm to solve the problem of overflying multiple interest points in a given area. \\
			\cline{2-3}
			&\cite{L_nonH_16}&The adaptive differential multi-objective
			optimization algorithm was used to find the optimal solution to avoid obstacles.\\
			\cline{2-3}
			&\cite{L_nonH_12}& Solved the problem of premature stagnation in genetic algorithm.\\
			\cline{2-3}
			&\cite{L_nonH_14}&Applied genetic algorithm and evolutionary robot to evolve neural network controller.\\
			\cline{2-3}
			&\cite{L_nonH_13}&Combined genetic algorithm and TSP algorithm to minimize the path length and adopted a new cellular decomposition to avoid concave obstacles.\\
			\cline{2-3}
			&\cite{L_nonH_11}&Applied Bayesian method to improve the genetic algorithm.\\

			\hline
		\end{tabular}
	\end{center}
\end{table*}

\subsubsection{Neural Network Algorithm}
Artificial neural network (ANN) can process different types of input data, such as 
position information from GPS, image information, and then output the next movement decision of UAV. 
\cite{L_nonH_6} proposed a path planning algorithm based on ANN, 
which directly processed the image of the visual system on the UAV, 
and applied the back-propagation neural network to get the best control command of the next movement (forward, right, left, {and so on}). 

In 3-D obstacle avoidance  
for UAV, the UAV's movements are constrained by a series of conditions. 
How to smooth the flight path of UAV to satisfy the flight restrictions of 
UAV, and how to evaluate the flight path and finally get the relatively
optimal path are the main challenges in the 3-D obstacle avoidance. 
\cite{L_nonH_8} proposed a path planning algorithm, 
where a Generalized Fuzzy Competitive Neural Network (G-FCNN) was applied to design the path. 
The irregular obstacles are distributed in 3-D environment,
which brings great difficulty for the design of obstacle avoidance algorithms.
{Considering avoiding random moving obstacles, DRL algorithm was applied} in \cite{L_nonH_1}, where neural 
network and Q-learning were used to learn from the experience.
Combined with dual network and priority sampling, the accuracy of collision avoidance will reach to $97.5\%$ after 13000 times of training.

{To solve obstacle avoidance problems indoors with dynamic obstacles, }an obstacle avoidance algorithm based on DRL was proposed to learn surrounding environment from monocular vision in \cite{L_nonH_2}, 
which used Conditional Generative Adversarial Network (CGAN) to build the depth map and finally used the depth map to make collision avoidance decisions. 
This method retained the key information about the environment during long observation,
{which can be applied to predict the motion information of objects, so as to better avoid dynamic obstacles such as human}. 
In \cite{L_nonH_3}, the depth map of monocular RGB image was predicted by using convolutional neural network (CNN), 
then the depth map is used to predict and avoid obstacles.
The size of image input by CNN network was shrunk to ensure real-time processing, 
and the mean filtering and histogram equalization were used to ensure the reliability of depth information. 
An algorithm combining the depth neural network and epipolar geometry was proposed to extract the depth value information of the image through deep learning in \cite{L_nonH_4}. 
Epipolar geometry was used to reduce the dependence on the training model, 
which improved the performance of the system 
when there are differences between the real data and the training data.

{CNN was applied in \cite{M_add_NN1} to estimate the optical flow of the pictures obtained by the camera. 
The decision of turning left or right is made by comparing the sum of the optical flow of left half-plane and right half-plane of the pictures.
Besides, the expansion of the optical flow is calculated to recognize wall-like frontal obstacles.
The experiment with a DJI F550 drone was carried out near a bike path, which verified the proposed obstacle avoidance algorithm.
}

\subsubsection{Reinforcement Learning}
Reinforcement learning is usually applied in the problem with continuous decision, which is often used in obstacle avoidance and path planing of UAV.

To enable the UAV to avoid obstacles when it is trapped, 
\cite{L_nonH_7} introduced the adaptive and random exploration (ARE) method in obstacle avoidance problem. 
This algorithm combines self-learning and random search, 
so that the UAV can avoid obstacles and simultaneously conquer learning mistakes.

{As to the obstacle avoidance algorithm of multiple UAVs, 
a multi-UAV anti-collision framework based on reinforcement learning was proposed in \cite{L_nonH_26}, so as to avoid collision with other UAVs. 
Four elements of environment control, action space, environment model, and return value 
in the decision-making process of UAV were analyzed in details, 
so as to effectively solve multi-UAV anti-collision problem.
To prevent collision with dynamic obstacles in the environment, \cite{L_nonH_5} proposed Experience-shared Advantaged Actor - Critic (ES-A2C) algorithm innovatively, where the UAV can share the learned experience with other UAVs in the swarm.
Compared with traditional algorithm such as Q learning algorithm and Actor-Critic algorithm, the ES-A2C algorithm has higher success rate and shorter training period.}

{In order to solve the collision avoidance problem between multiple UAVs, }\cite{L_nonH_27} proposed a multi-UAV obstacle avoidance method based on two-stage reinforcement learning, which firstly adopted the monitoring method with loss function, 
and then adopted the traditional reinforcement learning method in the second stage to overcome the shortcomings of the traditional reinforcement learning methods, 
such as high variance and poor repeatability.

{In order to ensure safety of UAV in high-dynamic multi-obstacle environment, \cite{M_add_27.5} proposed a distributed DRL framework, which applied LSTM to help UAVs capture more information from the dynamic environment. A clipped DRL loss function derived from UAV's exploration in the environment was proposed, which is more suitable for high-dynamic environment. Compared with DQN, Double DQN (DDQN), and other advanced DRL algorithms, this algorithm has better convergence property.}

\subsubsection{Genetic Algorithm}
Genetic algorithm is one of the intelligent optimization algorithms, 
which has been widely applied in UAV path planning. 
The genetic algorithm encodes the solutions of the optimization model into the chromosomes, 
evaluates the fitness of the chromosome, 
and retains the chromosome with high fitness to the next generation. 
This process is repeated until the optimized solution is found.

The collision avoidance problem of UAV can be converted into a path planing problem,
which is further equivalent to a Traveling Salesman Problem (TSP) \cite{L_nonH_9}.  
When solving TSP problem, 
genetic algorithm was applied for path planning and avoiding obstacles. 
\cite{L_nonH_10} first solved the problem of overflying multiple points of interest in a given area by Bellman Ford algorithm, 
and then used genetic algorithm to find the optimal path. 
\cite{L_nonH_16} defined the path
planing problem of UAV as a multi-objective optimization problem,
and proposed a new multi-gene structure to describe the
path, in which the adaptive adjustment, crossover, and mutation
strategies were adopted, and the adaptive differential multi-objective
optimization algorithm was applied to obtain the optimal
solution to avoid obstacles and meet the flight restrictions of UAV.
\cite{L_nonH_12} used the minimum spanning tree and adaptive tournament selection to quantify and control the genetic diversity, 
which solved the problem of premature stagnation.
For the obstacle avoidance problem in multi-UAV scenario, 
\cite{L_nonH_14} used genetic algorithm and evolutionary robot to evolve neural network controller, solving the obstacle avoidance problem of multi-UAV.

In agricultural scenarios, 
UAV path planning algorithm usually needs to maximize the coverage area to complete specific tasks such as pesticide spraying. 
Besides, the agricultural environment often consists of concave obstacles, which brings difficulties for collision avoidance algorithms.
\cite{L_nonH_13} combined genetic algorithm and TSP algorithm to minimize the path length and adopted a new cellular decomposition to avoid concave obstacles.
%In agricultural application scenarios, 
%there is often a need for path coverage. 
%Under the condition of path coverage, 
%genetic algorithm and TSP algorithm were combined 
%to minimize the path length, 
%and a new cellular decomposition was adopted 
%to avoid concave obstacles in \cite{L_nonH_13}. 
In order to avoid concave obstacles, 
\cite{L_nonH_11} applied Kalman filter to predict the state of obstacles, 
and then used Bayesian method to improve the genetic algorithm 
so that the UAVs will be easier to bypass concave obstacles.

\subsubsection{Simulated Annealing Algorithm}
%The simulated annealing algorithm can accept the solution that is less than the current one in a certain range, 
%which helps itself jump out of the local optimal solution 
%and get the global optimal solution, 
%so it can be well used to optimize UAV collision prevention and path planning.
\begin{table*}[!t]
	\label{Table_6}
	\caption{Summary of classic methods(2)}
	\begin{center}
		\begin{tabular}{|m{0.2\textwidth}<{\centering}|m{0.1\textwidth}<{\centering}|m{0.61\textwidth}|}
			\hline
			Category  & Reference & \makecell[c]{One-sentence summary}\\
			\hline
			\multirow{3}[12]{\linewidth}{\makecell[c]{Simulated annealing\\ algorithm (SAA)}}
			&\cite{L_nonH_20} &Adopted SAA to improve the global search efficiency of collision avoidance.\\
			\cline{2-3}
			&\cite{L_nonH_21} &Applied SAA to obtain the approximate optimal solution in a short period of time.\\
			\cline{2-3}
%			&\cite{L_nonH_22} &UFMS was proposed to optimize the target with local search strategy and simulated annealing algorithm.\\
%			\cline{2-3}
			&\cite{L_nonH_23}&Applied simulated annealing algorithm to get the optimal path under the largest area coverage in path planning problem under multiple constraints.\\
			\hline

			\multirow{4}[13]{\linewidth}{\makecell[c]{Ant Colony \\ Algorithm (ACA)}}
			&\cite{L_nonH_25}&Applied ACA to deal with the trajectory planning problem of UAV.\\
			\cline{2-3}
			&\cite{L_nonH_24}&Introduced a guidance factor of target node pheromone and re-excitation learning mechanism to improve the ACA in collision avoidance.\\
			\cline{2-3}
			&\cite{L_nonH_28}&Applied ACA to search for the shortest path for UAV in mountain environment.\\
			\cline{2-3}
			&\cite{L_nonH_29}&Proposed a new multiple ant colony algorithm.\\
			\hline

		\end{tabular}
	\end{center}
\end{table*}
Simulated annealing algorithm (SAA) has the characteristic of probabilistic jumping out of local optimum, 
which can accelerate the path planing. 
\cite{L_nonH_20} adopted SAA to improve the global search efficiency 
and introduced random mutation operation on the basis of genetic algorithm to improve the population diversity. 
\cite{L_nonH_21} defined the detection search target problem as a TSP problem. 
SAA was used to obtain the approximate optimal solution in a short time, 
which can effectively guide UAV escaping from threat areas. 
For the path planning problem of multiple UAVs, 
a new path planning method for UAV was proposed in \cite{L_nonH_23}, 
which considered three constraints including the number of UAVs, 
obstacle avoidance and coverage area. 
The flight area was decomposed into small areas, 
then K-means algorithm was applied to gather the target points. 
Finally, SAA was adopted to get the optimal path under the largest coverage area, 
solving the path planning problem under multiple constraints.

\begin{figure}[!t]
	\centering
	\includegraphics[width=0.4\textwidth]{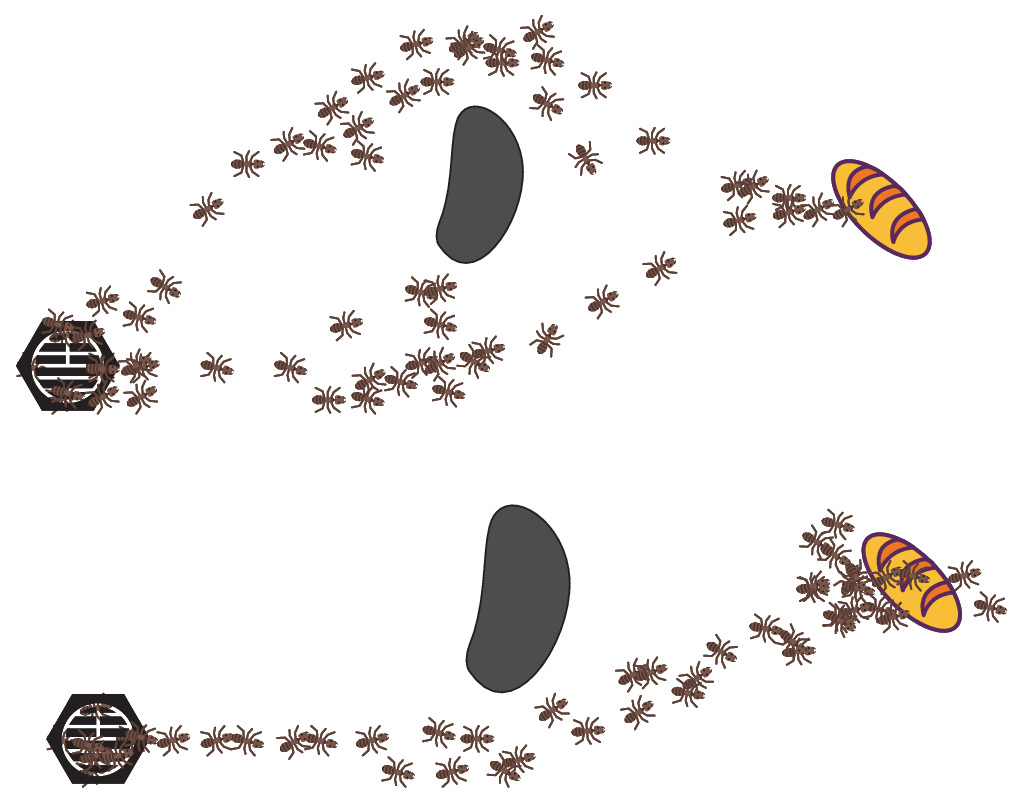}
	\caption{Ant colony algorithm}
	\label{Ant Colony Algorithm}
\end{figure}

\subsubsection{Ant Colony Algorithm}
As shown in Fig. \ref{Ant Colony Algorithm}, ant colony algorithm (ACA) is an algorithm that simulates the optimal path for ants to find food.  
When a better path is obtained, 
a positive feedback signal will be sent 
to guide the ants to the optimal path. 
In addition, the diversity of ant colony algorithm can avoid local optimization.

ACA was used to solve the trajectory planning problem of UAV in \cite{L_nonH_25}. 
The trajectory model of UAV is derived, 
and the simulation results showed that UAVs can accurately complete the path planning mission in a 3-D environment within 16 times of iteration.
\cite{L_nonH_24} introduced a guidance factor of target node pheromone in the process of the state transition of traditional ACA. 
Besides, re-excitation learning mechanism was applied to update pheromone on the path.
Those two improvements greatly increased the convergence speed.

\cite{L_nonH_28} proposed an ant colony based obstacle avoidance method 
for UAV in mountain environment. 
Firstly, Voronoi polygon was used to get the initial feasible solution of obstacle avoidance path in mountain environment, 
and then ACA was used to search for the shortest path. 
Finally, unnecessary obstacles were eliminated, 
which improved the searching speed and shortened the flight path.

For multiple UAVs, 
the multiple ant colony algorithm was proposed in \cite{L_nonH_29}. 
The concept of antibody similarity in immune optimization algorithm is introduced to measure the similarity of feasible solutions. 
For multiple similar individuals, only the representative individuals are selected for calculation, so as to improve the searching speed.

The pros and cons for collision avoidance methods are shown in Table \uppercase\expandafter{\romannumeral9}.

\begin{table*}[!t]
	\label{Table_CollisionAvoidance_PandC}
	\caption{Pros and cons of collision avoidance methods}
	\renewcommand{\arraystretch}{1.1} %looser
	\begin{center}
		\begin{tabular}{|m{0.13\textwidth}|m{0.1\textwidth}|m{0.24\textwidth}|m{0.24\textwidth}|m{0.17\textwidth}|}
			\hline
			\multicolumn{2}{|c|}{Name of the methods}& \multicolumn{1}{c|}{Pros}& \multicolumn{1}{c|}{Cons}& \multicolumn{1}{c|}{Common features} \\
			\hline
			\multirow{6}[19]{\linewidth}{Classic Collision Avoidance Algorithms} & \multirow{2}[0]{\linewidth}{Geometric Methods} & $\bullet$ Low computation overhead. &$\bullet$ Low accuracy when the number of obstacles is large. & \multirow{6}[19]{\linewidth}{$\bullet$ No iteration.\\$\bullet$ Fast calculation.\\$\bullet$ Low accuracy.}\\
			
			& & $\bullet$ Low cost.\vspace{0.5em}&$\bullet$ Large turning radius.\vspace{0.5em}&\\
			\cline{2-4}
			&\multirow{2}[5]{\linewidth}{Graph Theory Methods} & \multirow{2}[5]{\linewidth}{$\bullet$ Global optimization.}  & $\bullet$ High requirements for sensing since the global obstacle distribution is required.& \\
			&&&$\bullet$ Large construction time of the graph.&\\
			\cline{2-4}
			& \multirow{2}[-3]{\linewidth}{Artificial Potential Field} &\vspace{0.5em}$\bullet$ Low computation overhead.\vspace{0.5em}&
			\multirow{2}[-9]{\linewidth}{$\bullet$ Easy to be trapped into local minimum when the number of obstacles is large.}\vspace{1em}&\\
			
			&&$\bullet$ Low cost.\vspace{0.5em}&&\\
			\hline
			\multirow{10}[40]{\linewidth}{Heuristic Collision Avoidance Algorithms} 
			& \multirow{2}[0]{\linewidth}{Neural Network Algorithm} &
			$\bullet$ Able to process pure image, which decrease the calculation overhead of sensing.&$\bullet$ Prone to over fitting. &

			\multirow{10}[40]{\linewidth}{$\bullet$ Good generality.\\$\bullet$ Low calculation speed.\\$\bullet$ Easy to fall into local optimization.}\\
			&&$\bullet$ Good performance to solve nonlinear problems. &$\bullet$ Easy to fall into local optimization.&\\
			
			\cline{2-4}
			&\vspace{0.2em}Reinforce- ment Learning&$\bullet$ Global optimization.&$\bullet$ High variance and low reproducibility.&\\
			\cline{2-4}
			& \multirow{3}[5]{\linewidth}{Genetic Algorithm} &\multirow{3}[5]{\linewidth}{$\bullet$ Global optimization\vspace{0.5em}\\$\bullet$ Good performance on collision avoidance of UAV swarm}& $\bullet$ Premature convergence and stagnation problem.&\\
			&&& $\bullet$ High requirements for sequences coding. &\\
			&&& $\bullet$ Low efficiency.&\\
			\cline{2-4}
			&\multirow{2}[0]{\linewidth}{Simulated Annealing Algorithm} &\multirow{2}[7]{\linewidth}{$\bullet$ Probabilistic jumping out of local optimization.} & \vspace{1.5em}$\bullet$ Greatly affected by the initial value. &\\
			&&&$\bullet$ Slow convergence.\vspace{1.5em}&\\
			\cline{2-4}
			&\multirow{2}[7]{\linewidth}{Ant Colony Algorithm} &\multirow{2}[7]{\linewidth}{$\bullet$ Reduce the probability of local optimization because of population diversity.} & $\bullet$ Greatly affected by the initial value. &\\
			&&&$\bullet$ A contradiction between population diversity and convergence rate.&\\

			\hline
			
		\end{tabular}
	\end{center}
\end{table*}

\section{Future Trends}
In order to enhance the anti-collision capability of UAVs, most of current researches tend to improve the performance of collision avoidance algorithm. On the contrary, the performance of sensing has an impact on the performance of anti-collision capability of UAVs. Fast obstacle sensing techniques can shorten the sensing time and provide more response time during collision prediction and collision avoidance procedures. Fast wireless networking techniques enable UAV swarm to spread the sensing information quickly, and improve the cooperative sensing performance of UAV network. In this section, future trends of anti-collision techniques for UAVs are summarized from two aspects, i.e, fast obstacle sensing and fast wireless networking.
\subsection{Fast Obstacle Sensing}
\subsubsection{Joint Sensing and Communication Based Obstacle Sensing}
The obstacle sensing methods consists of cooperative and non-cooperative obstacle sensing methods. 
In the cooperative obstacle sensing methods, the communication between UAV and obstacle is applied to recognize and localize the obstacle. 
In the non-cooperative obstacle sensing methods, 
the active or passive sensors are applied to recognize and localize the obstacles. 
However, with joint sensing and communication (JSC) technology, 
the cooperative and non-cooperative obstacle sensing methods can work simultaneously for obstacle sensing. 
The communication signal is used for cooperative obstacle sensing, meanwhile, 
the communication signal can be applied for radar sensing, positioning, imaging, {and so on}, 
which belong to non-cooperative sensing. 
Hence, the JSC technology can realize fast and comprehensive sensing for obstacles, 
which is crucial for obstacle avoidance. The JSC technology actually attracts wide attention in the era of 6th generation mobile networks (6G) \cite{M_FT_1}.

In the JSC system, the performance of sensing and communication can be benefited from the joint design. 
The fuse of communication and sensing results improves the sensing accuracy \cite{M_FT_2}.
And the prior information of sensing can be applied to improve the performance of wireless networking \cite{M_FT_3}. Moreover, with JSC technology, the spectrum and hardware can be reused by sensing and communication, which improves the resource efficiency \cite{M_FT_5}. The application of JSC in UAV networks realizes the simultaneous cooperative and non-cooperative obstacle sensing, which has the potential to realize fast sensing and networking for UAV swarm and finally improve the obstacle avoidance capability.
The integration of positioning, communication, and radar is a promising technology for anti-collision system of UAVs, because it can realize comprehensive sensing and communication simultaneously. With the combination of FDA and OFDM technology, a FDA-OFDM scheme with broad application prospect was proposed in \cite{M_FT_7}, which realizes radar, communication, and positioning.
In recent years, the communications over mmWave and THz spectrum bands have attracted wide attention. With the wide bandwidth on mmWave and THz band, the radar imaging is promising to be deployed. Since the radar image can provide detailed information of the obstacles, it can be applied to recognize the obstacles. 
In \cite{M_FT_8}, the airborne MIMO radar and space-time coded (STC) waveform was designed, which is used to realize the joint communication and synthetic aperture radar (SAR) system. \cite{M_FT_10} proposed a joint waveform for simultaneous communicate and synthetic aperture radar imaging.
\subsubsection{AI Chip Based Obstacle Sensing}
The machine learning algorithms are widely applied in obstacle sensing. With the chip-level obstacle sensing, the speed and accuracy of machine learning based obstacle sensing are further improved. The machine learning algorithms can be realized by central processing unit (CPU), graphics processing unit (GPU), field programmable gate array (FPGA) and application specific integrated circuit (ASIC). Generally, the FPGA has better performance/Watt compared with CPU and GPU \cite{M_FT_11, M_FT_12}, and CPU is more versatile. 
However, the high overhead suffered from the operating system (OS) will degrade the speed of computation \cite{M_FT_13}. The GPU has high speed of computation. However, the power consumption is high. The FPGA and ASIC have low power consumption compared with CPU and GPU, which are suitable to be implemented in UAVs. However, the versatility of FPGA and ASIC is low. Besides, the ASIC is customized, providing higher energy efficiency and speed of computation compared with FPGA, which has potential to be applied in small-size UAVs.

In \cite{M_FT_14}, the UAV is installed with the AI chip, whose structure is like the nervous system of fruit fly. And \cite{M_FT_14} verified that the UAV can avoid obstacles with relatively low energy consumption. In \cite{M_FT_15}, with NVIDIA Jetson TX2 implemented in UAV, the data of six 4K cameras is processed for obstacle sensing. A chip named “Navion” is developed in \cite{M_FT_16} for UAV, with tiny energy consumption, the chip can process 171 frames per second for positioning. Overall, for the small UAVs, the ASIC or FPGA is an optimal choice. However, for the large UAVs, the GPU can be applied. 
In academia and industry, the implementation of AI chip on UAVs for obstacle sensing has attracted wide attention to enhance the capability of obstacle sensing for UAVs.

\subsection{Fast Wireless Networking} 
The fast wireless networking is of vital importance to enhance the anti-collision performance of UAVs. The sensing information of obstacles with fast wireless networking can be spread to other UAVs in the network quickly, such that UAVs can make anti-collision decision in advance and improve the anti-collision performance. The research of wireless networking of UAVs mainly focuses on neighbor discovery, multiple access control and the routing scheme.

\subsubsection{Neighbor Discovery}
The prior information of the distribution of neighbors can accelerate the speed of neighbor discovery.  
Compared the neighbor discovery algorithms without the prior information of radar sensing, the time of neighbor discovery algorithm is significantly reduced using radar sensing information \cite{M_FT_17}. 
\cite{M_FT_18} proposed a three handshake neighbor discovery algorithm for radar-communication integrated system (RCIS) network.
The algorithm applies radar sensing to obtain the location information of neighbors, 
so as to increase the speed of neighbor discovery.
In \cite{M_FT_19},  the efficiency of neighbor discovery was improved using dual phased array radar.
\cite{M_FT_20} proposed a neighbor discovery algorithm based on joint radar and communication (JRC) technique.
The algorithm takes advantage of the neighbors’ location information obtained from radar sensing, and avoids repeated attempts on directions without neighbors. The analysis and simulation shows that the algorithm can greatly improve the neighbor discovery speed.

\subsubsection{Multiple Access Control}
In order to design the fast multiple access control (MAC) scheme, UAVs can estimate the number of competing nodes in the network with sensing information, thereby reducing the congestion probability of the MAC protocol.
As the number of nodes increases, the access competition of nodes in a single channel will become more and more intense. 
By applying multi-channel mechanism, the load and the delay of the network will be reduced. 
\cite{M_FT_21} proposed a MAC protocol applying integrated sensing and communication (ISAC) technique, where radar and communication can cooperate with each other.
The simulation result shows that MAC protocol with ISAC can achieve high throughput.
Feng \emph{et al.} designed a MAC protocol under multi-channel opportunistic reservation based on cognitive radio (CogMOR-MAC) \cite{M_FT_22}.
The CogMOR-MAC applies multi-channel opportunistic reservation mechanism to reserve resources, which reduces the access {conflict} and waiting time of UAV nodes.
\cite{M_FT_23} proposed a multi-channel MAC protocol with directional antennas  (MMAC-DA) for UAV nodes to transmit and receive directionally.
MMAC-DA applies common control channel to negotiate control information, 
which increases the throughput of UAV network.

\subsubsection{Routing}
The assistance of sensing information is promising to improve the efficiency of the routing scheme.
\cite{M_FT_24} proposed a fountain-code based greedy queue and location information assisted routing protocol, which applies velocity and distance information to improve routing efficiency. 
Aiming at solving the performance degradation caused by flooding in routing protocol, \cite{M_FT_25} proposed a location information and cluster-based routing protocol, where each node selects the node with the largest forwarding distance as the next hop to reduce the number of hops. 
In order to reduce the routing overhead, \cite{M_FT_26} proposed a location-assisted zone routing protocol, which has better performance on routing overhead and end-to-end delay compared with traditional routing schemes.

 \section{Conclusion}

{Due to the complex and changeable flight environment, flexible and reliable anti-collision technologies are urgently needed to ensure UAV's safety.
To meet such requirement, we provide an overview for UAV anti-collision technologies in this article.}
Firstly, we introduced laws and regulations of governments on UAV flight safety.
{The laws and regulations establish a bridge between academia and application.}
Secondly, anti-collision algorithms were introduced according to three aspects, namely, obstacle sensing, collision prediction, and collision avoidance.
{We select the representative articles of anti-collision algorithms and classified them in details, so as to provide a clear structure for readers.}
Finally, the future trends of UAV's anti-collision techniques are revealed from two aspects, i.e, fast obstacle sensing
and fast wireless networking. 
{Combined with state-of-art technologies of sensing, computation and communication, we provide a new perspective for UAV anti-collision technologies.}
This article may provide guidelines for the design of anti-collision mechanisms for UAVs.

\end{spacing}
\end{document}